\documentclass[conference]{IEEEtran}
\IEEEoverridecommandlockouts
\usepackage{cite}

\usepackage{amsmath,amssymb,amsfonts}
\usepackage{algorithm}
\usepackage{algorithmicx}
\usepackage{algpseudocode}

\usepackage{graphicx}
\usepackage{xspace}
\usepackage{tabularx}
\usepackage{multirow}
\usepackage{textcomp}
\usepackage{xcolor}
\def\BibTeX{{\rm B\kern-.05em{\sc i\kern-.025em b}\kern-.08em
    T\kern-.1667em\lower.7ex\hbox{E}\kern-.125emX}}

\newcolumntype{b}{X}
\newcolumntype{s}{>{\hsize=.45\hsize}X}
\newcolumntype{x}{>{\hsize=.25\hsize}X}

\usepackage{caption}

\usepackage{xcolor,colortbl}
\definecolor{grey}{rgb}{0.89, 0.898, 0.906}

\begin{document}

\title{Risks, Causes, and Mitigations of Widespread Deployments of Large Language Models (LLMs): A Survey\
}

\author{
    \IEEEauthorblockN{Md Nazmus Sakib}
    \IEEEauthorblockA{
        \textit{\small{Dept. of CSE}} \\
        \textit{Pabna University of Science}\\
        \textit{and Technology} \\
        Pabna, Bangladesh \\
        \footnotesize{nazmus.200103@s.pust.ac.bd}
    }
    \and
    \IEEEauthorblockN{Md Athikul Islam}
    \IEEEauthorblockA{
        \textit{\small{Dept. of Computer Science}} \\
        \textit{Boise State University} \\
        Boise, ID, USA \\
        \footnotesize{mdathikulislam@u.boisestate.edu}
    }
    \and
    \IEEEauthorblockN{Royal Pathak}
    \IEEEauthorblockA{
        \textit{\small{Dept. of Computer Science}} \\
        \textit{Boise State University} \\
        Boise, ID, USA \\
        \footnotesize{royalpathak@u.boisestate.edu}
    }
    \and
    \IEEEauthorblockN{Md Mashrur Arifin}
    \IEEEauthorblockA{
        \textit{\small{Dept. of Computer Science}} \\
        \textit{Boise State University} \\
        Boise, ID, USA\\
        \footnotesize{mdmashrurarifin@u.boisestate.edu}
    }
}

\maketitle

\begin{abstract}
Recent advancements in Large Language Models (LLMs), such as ChatGPT and LLaMA, have significantly transformed Natural Language Processing (NLP) with their outstanding abilities in text generation, summarization, and classification. Nevertheless, their widespread adoption introduces numerous challenges, including issues related to academic integrity, copyright, environmental impacts, and ethical considerations such as data bias, fairness, and privacy. The rapid evolution of LLMs also raises concerns regarding the reliability and generalizability of their evaluations. This paper offers a comprehensive survey of the literature on these subjects, systematically gathered and synthesized from Google Scholar. Our study provides an in-depth analysis of the risks associated with specific LLMs, identifying sub-risks, their causes, and potential solutions. Furthermore, we explore the broader challenges related to LLMs, detailing their causes and proposing mitigation strategies. Through this literature analysis, our survey aims to deepen the understanding of the implications and complexities surrounding these powerful models.

\end{abstract}


\section{Introduction}
\label{sec:Introduction}
LLMs are pre-trained on extensive corpora with vast numbers of parameters, excelling in various NLP tasks such as text generation, summarization, classification, machine translation, and question answering \cite{chalkidis-etal-2019-large, Chu_Song_Yang_2024, guo-etal-2022-auto, huang2020reducing}. In 2023, several major LLMs were released, including OpenAI's ChatGPT \cite{openai2024gpt4}, Meta AI's LLaMA \cite{touvron2023llama}, and Databricks' Dolly 2.0. These models exemplify the trend toward exponentially increasing parameters, such as GPT-2 with 1.5 billion parameters and GPT-3 with 175 billion parameters \cite{floridi2020gpt}. The deployment of these models spans applications in chat agents, computational biology, programming, creative domains, knowledge work, medicine, reasoning, robotics, and the social sciences \cite{kaddour2023challenges}.

% bommasani2022opportunities

Despite the remarkable success of LLMs, they pose several challenges that are unprecedented for humans \cite{kaddour2023challenges}. Diverse organizations are deploying apps that integrate LLMs while existing apps and features are constantly being updated with these new LLMs \cite{rillig2023risks}. However, these rapid updates raise various concerns, including academic integrity, copyright issues, and environmental impacts \cite{rillig2023risks}. Moreover, as LLMs grow in size, their insatiable demand for data becomes apparent. These models are now trained on such vast amounts of data that humans can no longer manually scrutinize it all \cite{kaplan2020scaling}. In addition, evaluation results may be flawed because the training data could include instances from the test data \cite{kaddour2023challenges}. This large-scale pre-training also introduces issues like bias and fairness, as well as ethical concerns. 

When a new language model is introduced, researchers often investigate its challenges and limitations \cite{TrusWhalMouz2023m0, pan_privacy_risks}. Additionally, some studies focus on identifying and mitigating specific risks associated with these models \cite{TrusWhalMouz2023m0}. As language models rapidly evolve, there is an increasing need for comprehensive literature that addresses these areas concurrently. Surveys or reviews that cover the issues, solutions, and underlying causes related to language models are essential. While existing surveys may address general AI risks and solutions \cite{park2023ai}, the security and privacy aspects of language models \cite{YAO2024100211, pan_privacy_risks}, or the challenges and solutions of specific models like ChatGPT \cite{Y_Wang_A_Survey_2023, gupta_from_chatgpt_2023}, a more diverse survey is needed. Such a survey would systematically outline the risks, causes, and mitigations associated with individual language models. To fill this gap, we propose a comprehensive survey to identify risks posed by specific language models, explore the reasons behind these risks, and suggest potential mitigation techniques.

Our survey analyzes and discusses the risks, causes, and mitigations associated with the widespread deployment of LLMs. The survey begins by listing the major risks associated with specific LLMs, followed by identifying sub-risks through synthesized data collected from relevant literature. For each sub-risk, the survey identifies which LLMs are affected, defines the root causes, and provides possible mitigations. All explanations are supported by the papers extracted from the literature search. In the final stage, the survey lists the general reasons behind the risks associated with LLMs and discusses mitigation techniques to address these causes.

The paper is organized into the following sections. Section \ref{sec:related_work} reviews previous survey work on the challenges, causes, and solutions related to Artificial Intelligence (AI), LLMs, or specific LLMs. Section \ref{sec:methodology} discusses methodology. Section \ref{sec:risks} examines the risks and sub-risks associated with specific LLMs. Section \ref{sec:causes_of_risks} explores the root causes of these risks, and Section \ref{sec:mitigations} provides mitigation techniques to address these risks. Finally, Section \ref{sec:conclusion} summarizes the key takeaways from the paper.

\section{Related Work}
\label{sec:related_work}
There are three categories of related work concerning challenges and solutions, which include surveys on AI or LLMs in general, as well as those focusing on a specific LLM.

The first category of surveys addresses risks and potential solutions in AI and machine learning (ML) \cite{park2023ai, tian_a_comprehensive_2022}. Park et al. conducted a survey detailing various AI deceptions, including fraud, election tampering, manipulation, and feints, and suggested mitigation techniques like risk assessment, documentation, record-keeping, transparency, and human oversight \cite{park2023ai}. In contrast, Tian et al. surveyed different poisoning attack strategies such as label and data manipulation, along with corresponding countermeasures, including data-driven and model-driven approaches \cite{tian_a_comprehensive_2022}.

The second category of surveys focuses on evaluating risks and solutions specific to LLMs in general. Yao et al. conducted a study on security and privacy issues related to LLMs, identifying their positive contributions to code security, data security, and privacy, as well as their involvement in various attacks targeting hardware, operating systems, software, networks, and users \cite{YAO2024100211}. Similarly, Dong et al. conducted a survey covering a wide range of attacks on LLMs during both training and inference phases, along with discussions on defense strategies \cite{dong2024attacks}. Hadi et al. set out with the intention of surveying applications, challenges, limitations, and future prospects but primarily provided an overview of generative AI and LLMs, focusing on their tasks and applications rather than comprehensively covering all aspects \cite{hadi2023large}. 

The final category involves surveys focused on individual LLMs, with practitioners primarily examining ChatGPT due to its recent success. Wang et al. elucidated the evolution of AI-generated content (AIGC) over time, particularly with the latest introduction of pre-trained large models \cite{Y_Wang_A_Survey_2023}. Their study delved into various applications of AIGC while also shedding light on security and privacy threats that pose risks to AIGC services, with ChatGPT serving as a central reference point. Similarly, Yang et al. conducted a survey specifically on ChatGPT \cite{yang_harnessing_2024}. They began by constructing an evolutionary tree for LLMs and proceeded to offer a brief introduction to popular models like BERT and ChatGPT. Additionally, they explored key considerations from a data perspective, concluding their discussion with an examination of ChatGPT's efficiency and trustworthiness.

Our survey distinguishes itself from previous literature by focusing on the unique risks posed by individual LLMs, uncovering their root causes, and proposing viable solutions.

\section{Methodology}
\label{sec:methodology}
The methodology outlines detailed steps for conducting the survey. These steps encompass searching for relevant literature, applying filtering criteria, collecting and extracting data, and synthesizing and analyzing the findings.

\subsection{Searching Relevant Literature}

We employed Google Scholar to search for existing literature pertaining to the risks, causes, and mitigations associated with LLMs. The search strategy was focused on literature containing keywords such as "LLMs", "large language models", "risk assessment", and "risk mitigation".

\subsection{Filtering Criteria}

We included literature on LLMs that is relevant to fields such as NLP, security, artificial intelligence, privacy, and specific language models mentioned in the literature \cite{yang_harnessing_2024}. We excluded papers published as tutorials, presentations, comments, discussions, and keynotes. Additionally, we considered literature published from 2000 to 2024.

\subsection{Data Collection}

We adopted the data collection approach outlined by Kitchenham and Charters \cite{keele2007guidelines}, adhering to specific quality criteria for selecting literature:

\begin{itemize}
  \item Assessment of the retention and presentation of contexts and data sources.
  \item Evaluation of the clarity and coherence of reporting.
  \item Examination of the attention given to ethical considerations.
\end{itemize}

The first author conducted a thorough analysis of the titles and abstracts of each search result retrieved from Google Scholar, identifying papers deemed relevant for further review. Subsequently, all authors collectively reviewed the selected literature in full-text, applying filtering and quality criteria, resulting in the final inclusion of 47 papers for the study.

\subsection{Synthesis and Analysis}

The synthesis of this study involved extracting summarized findings regarding the risks, causes, and solutions associated with LLMs. To document the risks, we developed a table and manually identified their corresponding reasons. Additionally, we created another table to categorize the generic causes. These tables were later presented in the paper as both tabular representations and lists.

\section{Risks Associated with LLMs}
\label{sec:risks}
\begin{table*}[ht!]
\centering
\def\arraystretch{1.25}
\resizebox{\textwidth}{!}{\begin{tabular}{p{2.5cm}|p{2.5cm}|p{2.5cm}|p{2.5cm}|p{3cm}|p{1.5cm}}
\hline
\hline
Risks & Sub-Risks & LLMs Associated with Risks & Causes & Possible Mitigations & References \\
\hline
1. Privacy Issues & Exposing user data & GPT-Neo, GPT-3 & Memorizing training data & Curation and change in distribution of training data & \cite{carlini2023quantifying, zeng2024good, yan2024protecting, Zhou_Xiang_Chen_Su_2024} \\
\cline{2-6}
& Leakage of retrieval and training data & Llama-7b-Chat, GPT-3.5-turbo & Memorizing retrieval and training data in RAG & Privacy-preserving prompt tuning & \cite{zeng2024good, li2023privacypreserving}
\\
\cline{2-6}
& Revealing user activity & Bard & Using activity data for training & Opt-out option, consent form & \cite{gupta_from_chatgpt_2023} \\
\hline
2. Susceptibility to Adversarial Attacks & Harmful content generation & GPT & Prompt injection & Filtering out retrieved information & \cite{WU2024_unveiling_security_privacy}
\\
\cline{2-6}
& Vulnerable sentence embeddings & GPT, BERT & Capture of sensitive information in embeddings & Rounding, privacy-preserving mapping, subspace projection & \cite{pan_privacy_risks}
\\
\cline{2-6}
& Stealing API service & BERT & Easy imitation of victim model & Softening predictions, prediction perturbation & \cite{he-etal-2021-model}
\\
\hline
3. Ethical Concerns & Lack of reliability, trustworthiness, and accountability & BERT, RoBERTa, Gemma-7b, Llama-2-7b & Adversarial attacks, overfitting & Regularization, adversarial training, random smoothing & \cite{jin2020textfooler, li-etal-2020-bert-attack, le-etal-2022-shield, liu-etal-2022-flooding, fujiwarameasuring}
\\
\cline{2-6}
& High pricing & T5, BERT, GPT, and others & Higher energy consumption & Lighter and reduced parameterized models, faster hardware & \cite{liu2024green}
\\
\hline
4. Bias and Fairness & Social and environmental bias & GPT-4, Claude-2, Llama-2-70b, Zephyr-7b & Biased training data and model architecture & Hyperparameter tuning, instruction guiding, debias tuning & \cite{Elrod_2024, dong2024disclosure}
\\
\cline{2-6}
& Human-like biases and stereotypes & BERT, ELMo, GPT, GPT-2, RoBERTa, DeBERTa, T5 & Large biased human-written training corpora & Debiasing losses, auto-debias, prompt engineering, model fine-tuning & \cite{guo-etal-2022-auto, cmsf2022003003, tal2022fewer, huang2020reducing, bender_on_the_dangers_2021, dong2024disclosure, lin2024investigating}
\\
\hline
5. Adverse Environmental Effects & Financial instability and high CO2 emission & All LLMs & High energy consumption & Lighter and reduced parameterized models, faster hardware & \cite{liu2024green, shi2024efficient, bender_on_the_dangers_2021}
\\
\hline
6. Violating Legal or Regulatory Requirements & Possibility of using copyrighted data & Proprietary LLMs & Copies or close variations of copyrighted data used in training & Copyright regression, softmax regression & \cite{Chu_Song_Yang_2024, yan2024protecting}
\\
\hline
7. Disruption in Human Life & Health and wellness & All LLMs including GPT-3 & Huge text generation, low-quality scientific literature & Developing policies & \cite{de2023chatgpt}
\\
\cline{2-6}
& Financial instability & All LLMs & High efficiency of LLMs & Strong policymaking & \cite{eloundou2023gpts}
\\
\hline
\hline
\end{tabular}}
\caption{ Ablation study findings evaluated using three.}
\label{table:ablation-study}
\end{table*} 

\subsection{Privacy Issues}

Models that memorize training data excessively are prone to overfitting and can compromise user privacy. Notably, large models like GPT-Neo tend to retain significant amounts of training data, often resulting in repeated patterns and increased risk of privacy breaches \cite{carlini2023quantifying}. The retrieval and training dataset database in Retrieval-Augmented Generation (RAG) models such as Llama-7b-Chat and GPT3.5-turbo can expose private data, further increasing privacy risks \cite{zeng2024good}. Models like Bard use user activity data to train their models along with the original training data, which can lead to a tendency to reveal user activity information  \cite{gupta_from_chatgpt_2023}. 

\subsection{Susceptibility to Adversarial Attacks}

GPT models, particularly ChatGPT, are susceptible to various security vulnerabilities. For example, an adversary could potentially instruct ChatGPT to generate text that is harmful to society, create malware code, or even distribute malicious code libraries \cite{WU2024_unveiling_security_privacy}. Additionally, language models like BERT, GPT, or GPT-2 generate sentence embeddings that attackers could reverse-engineer, potentially exposing sensitive information \cite{pan_privacy_risks}. Moreover, fine-tuned publicly available BERT model APIs let attackers extract a local copy of a target BERT model, giving them a way to generate adversarial attacks against the original model. \cite{he-etal-2021-model}.

\subsection{Ethical Concerns}

Pre-training LLMs with GPUs demands substantial RAM usage, leading to higher costs for companies \cite{liu2024green}. Additionally, without proper regulations, LLM service providers may charge users premium prices. Importantly, LLMs should produce consistent outputs for texts with the same semantic meaning. However, their variable responses in these situations raise ethical concerns, including issues of reliability, trustworthiness, and accountability. Models like BERT and RoBERTa have demonstrated these vulnerabilities \cite{jin2020textfooler, li-etal-2020-bert-attack, le-etal-2022-shield}.

\subsection{Bias and Fairness}

Due to the training on massive datasets, certain LLMs, like GPT-4, Claude-2, Llama-2-70b,  Zephyr-7b exhibit bias toward current social and environmental topics, suggesting that they are significantly influenced by contemporary socio-political discourse \cite{Elrod_2024, agiza2024analyzing}. Moreover, Masked Language Modeling (MLM) of the larger version of BERT, RoBERTa, DeBERTa, T5 shows sentiment and human biases for male and female gender \cite{guo-etal-2022-auto, cmsf2022003003, tal2022fewer, bender_on_the_dangers_2021}.  Sentiment bias is another concern for models like BERT, which often display a significant degree of bias in how they interpret and generate text \cite{huang2020reducing}. GPT-3, trained on Common Crawl datasets, may produce sentences with high toxicity even when the prompt sentences are non-toxic \cite{bender_on_the_dangers_2021}.

\subsection{Adverse Environmental Effects}

While LLMs have been successful, they also bring certain adverse environmental effects, such as high energy consumption and contributing to digital divides \cite{rillig2023risks, bender_on_the_dangers_2021}. For example, LLMs like T5 and BERT consume large amounts of energy, resulting in significantly higher CO2 emissions \cite{liu2024green, shi2024efficient}. Training a BERT model, even without hyperparameter tuning, requires an amount of energy comparable to that of a trans-American flight \cite{bender_on_the_dangers_2021}.

\subsection{Violating Legal or Regulatory Requirements}

LLMs trained on extensive datasets can generate outputs that violate legal or regulatory requirements or closely resemble copyrighted material \cite{Chu_Song_Yang_2024}. These models typically utilize internet-sourced data, publicly available datasets, and occasionally proprietary information \cite{yan2024protecting}. As LLMs scale rapidly, it becomes increasingly challenging to ensure proprietary data is not used without authorization. Regulating these models is also complex. Another risk is the inadvertent exposure of private data when interacting with LLMs; for instance, Samsung Electronics disclosed sensitive information to ChatGPT on multiple occasions \cite{yan2024protecting}.

\subsection{Disruption in Human Life}

The substantial volume of text generated by LLMs can lead to misuse in the medical field, potentially posing a public health risk \cite{de2023chatgpt}. Additionally, this overwhelming influx of text may contribute to information overload and anxiety. LLMs like ChatGPT can also produce low-quality scientific literature, which might have adverse effects on human health \cite{de2023chatgpt}. The advanced capabilities and automation provided by LLMs have put many human jobs at risk, with nearly 19\% of roles experiencing at least 50\% of tasks coverage by LLMs \cite{eloundou2023gpts}. This could lead to significant economic disruption.

\section{Causes of Risks}
\label{sec:causes_of_risks}
The following are the general causes behind the risks associated with LLMs.

\subsection{Excessive Memorization of Training Data}

Deep language models are prone to memorizing training data, leading to overfitting \cite{liu-etal-2022-flooding, pmlr-v119-ishida20a, takeoka-etal-2021-low}. This memorization often results in the leakage of private data. Liu et al. demonstrated how BERT models suffer from poor generalization due to memorizing training data \cite{liu-etal-2022-flooding}. Similarly, Zhou et al. observed a memorization tendency in the GPT-Neo model, while models like OPT and Llama exhibited less propensity for memorization \cite{Zhou_Xiang_Chen_Su_2024}. Another finding is that bigger models memorize more than the smaller models \cite{schwarzschild2024rethinking}.

\subsection{Inherent Complexity of LLMs}

With the continually increasing size of LLMs and their capacity to perform tasks resembling human abilities, understanding them has become increasingly complex. This inherent complexity has hindered the utilization of LLMs in scientific research and data analysis \cite{singh2024rethinking}. TripoSR and GemMoE-Beta-1 models have demonstrated greater transparency in their reasoning compared to Gemma-7b and Llama-2-7b models \cite{fujiwarameasuring}. 

\subsection{Lack of Awareness of LLMs}

The end-users, policymakers, various stakeholders, and even the developers themselves may lack a thorough understanding of the serious risks posed by LLMs. For example, in the context of public health, it's critical to recognize that the CareCall chatbot occasionally makes promises akin to human capabilities, despite its inability to fulfill them \cite{10.1145/3544548.3581503}. Allowing such behavior could have severe consequences for businesses. 

\subsection{Testing and Evaluation Flaws}

Occasionally, LLMs are trained using the development and test sets of benchmark datasets, leading to improper evaluation, a phenomenon known as benchmark leakage \cite{zhou2023dont}. This issue has raised concerns regarding the fairness and reliability of LLM testing. Notably, models like OpenLLaMA-3B and LLaMA-2-7B exhibit adverse effects on evaluation due to benchmark leakage \cite{zhou2023dont}.

\subsection{Evolving Threat Landscape}

The threat landscape surrounding LLMs is evolving rapidly alongside their expansion. One such security concern is the emergence of "jailbreak prompts," which bypass the security measures of LLMs, compelling them to produce harmful content. Recent iterations of these prompts have demonstrated alarming success rates, with some achieving up to 99\% attack success rates (ASR) on the latest models like ChatGPT (GPT-3.5) and GPT-4 \cite{shen2023do}. Another emerging attack vector is known as "Indirect Prompt Injection," which coerces LLM-integrated applications into delivering intended adversarial content to end users. For instance, Bing Chat, operating on the GPT-4 model, has exhibited vulnerability to this attack \cite{Greshake_not_what_2023}.

\subsection{Lack of Strong Policy Making}

Insufficient policies, particularly in areas like data protection and security, can leave LLMs and their users on risk. For instance, third-party providers of LLMs may collect user data without obtaining proper consent or providing clear explanations regarding data usage \cite{yan2024protecting}. This leads to a strong privacy breach.

\subsection{Security Vulnerabilities}

The security vulnerabilities inherent in LLMs make them susceptible to manipulation in output generation. These vulnerabilities can be exploited to create fake news, spam emails, and other deceptive content \cite{Esmradi_a_comprehensive_2024}. Adversarial attacks targeting LLMs encompass various strategies such as model theft, aimed at extracting model shapes and parameters, data construction for mimicking training data, data poisoning to introduce malicious data, and model hijacking to perform unauthorized tasks \cite{Esmradi_a_comprehensive_2024}. Additionally, attacks on LLM applications include prompt injection, which leads to inconsistent outputs, and privacy leakage attacks \cite{Esmradi_a_comprehensive_2024}. For example: ChatGPT and Azure OpenAI (GPT-3.5 turbo) are vulnerable to prompt injection attack.

\subsection{Poor Data Quality}

Given that LLMs rely on pre-training with large datasets, it becomes crucial to ensure the quality of these datasets. Pre-processing and curating such vast datasets pose significant challenges \cite{hadi2023large}. If datasets inherently contain biases, cultural norms, and stereotypes, training LLMs on such data propagates these limitations throughout the models \cite{agiza2024analyzing}. Agiza et al. demonstrated how ideological and political biases can be ingrained in the Mistral-7b-v0.2 model \cite{agiza2024analyzing}.

\section{Mitigation Strategies}
\label{sec:mitigations}
\begin{figure}[h]
  \centering
  \includegraphics[width=0.8\columnwidth]{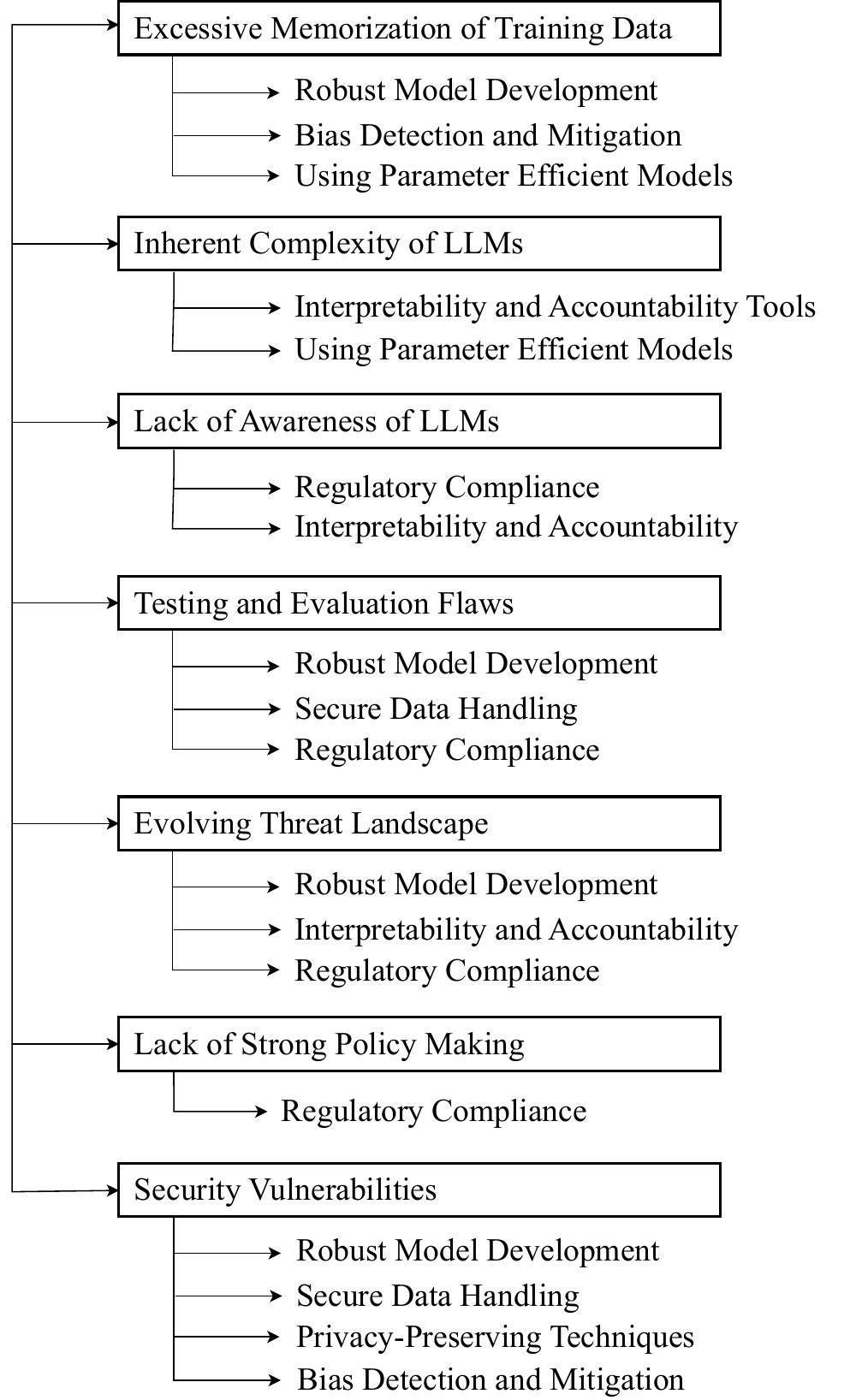}
  \caption{The boxes in the tree diagram represent the root causes of LLM risks, and the lists under each cause mention the mitigation techniques for that specific cause.}
  \label{fig:mitigation_strategies}
\end{figure}

The following are the mitigation techniques for addressing the underlying risks of LLMs. Figure \ref{fig:mitigation_strategies} illustrates these mitigation techniques for the various risk causes associated with LLMs.

\subsection{Robust Model Development}

LLMs require thorough development, involving extensive testing and evaluation processes to address security vulnerabilities and biases. Various techniques exist to mitigate issues like overfitting, including regularization, dropout, batch normalization, and label smoothing \cite{pmlr-v119-ishida20a}. Adherence to industrial standard guidelines and best practices is also essential for mitigating adversarial attacks. Moreover, adversarial training and ensemble methods are also widely used techniques for preventing adversarial attacks \cite{zeng-etal-2023-certified, wang2021infobert}.

\subsection{Privacy-Preserving Techniques}

There are various techniques available for preserving data privacy. One approach involves centralized privacy settings, where the service provider configures privacy settings on behalf of end-users \cite{shi2022just, li2022large}. Conversely, other methods empower end-users to set up privacy measures for their data themselves. An example of this is Privacy-Preserving Prompt Tuning (RAPT) \cite{li2023privacypreserving}.

\subsection{Regulatory Compliance}

With the ever-expanding size of language models and AI, it is crucial to establish robust regulatory compliance measures. Addressing compliance challenges involves ensuring data privacy and security, mitigating bias, promoting fairness, and enhancing transparency \cite{hubert2024regulatory}. Developing comprehensive governance frameworks is essential for effectively tackling these issues \cite{hubert2024regulatory}. Ethical language model development must be prioritized to safeguard against biases, promote fairness, and uphold accountability \cite{hubert2024regulatory}. Policies should recognize the broad functionalities and constraints of today’s LLMs, advocating for transparency, responsibility, and ethical application \cite{fujiwarameasuring}. Continuous monitoring is indispensable to promptly identify and rectify compliance issues. Establishing ethical guidelines and governance frameworks ensures that LLMs align with societal values and democratic principles \cite{agiza2024analyzing}. Additionally, Chu et al. proposed a softmax regression approach to help models avoid generating copyrighted data during training and inference \cite{Chu_Song_Yang_2024}.

\subsection{Secure Data Handling}

Following industry best practices, such as encryption and access control, is crucial to safeguard data from unauthorized access. Implementing strong encryption protocols ensures the secure storage and transmission of private or sensitive information. Additionally, when interacting with end-users and managing their data, it is vital to have effective consent management procedures in place to transparently communicate how data will be collected and processed \cite{hubert2024regulatory}.

\subsection{Bias Detection and Mitigation}

Fleisig et al. proposed an adversarial learning approach, while Dong et al. employed a probing framework with conditional generation to identify and address gender bias \cite{fleisig2022mitigating, dong2024disclosure}. Other techniques for mitigating bias include pre-processing, data filtering, prompt modification, and fine-tuning \cite{lin2024investigating}. For instance, GPT-3.5-turbo can undergo further debiasing through fine-tuning \cite{lin2024investigating}. Additionally, Huang et al. utilized Few-shot learning and Chain-of-Thought (CoT) methods for debiasing in code generation \cite{huang2024bias}.

\subsection{Interpretability and Accountability}

A highly interpretable LLM is likely to be more acceptable to end users \cite{fujiwarameasuring}. The fields of medicine and science necessitate highly interpretable LLMs to ensure their effective utilization. LLMs interpretability can be categorized into two methods: local and global. Local interpretability focuses on explaining a single output, whereas global interpretability aims to elucidate the LLM as a whole \cite{singh2024rethinking}. Local methods, such as perturbation-based methods, gradient-based methods, and linear approximations, are utilized to compute feature importance. Additionally, computing Shapley values represents a unique attribution method for LLMs \cite{frye2021shapley}. On the other hand, global explainability methods include probing and understanding the distribution of training data \cite{singh2024rethinking}. 

\subsection{Using Parameter Efficient Models}

Larger models often tend to memorize training data more extensively than their compact counterparts, making the latter preferable in certain scenarios \cite{carlini2023quantifying}. For instance, DistilBERT, a significantly streamlined version of the BERT model with a 40\% reduction in parameters, demonstrates robust performance across various NLP tasks compared to its parent BERT \cite{liu2024green}. Notably, DistilBERT retains 97\% of BERT's understanding capabilities while offering substantially faster inference times \cite{liu2024green}. Smaller models like DistilBERT are easier to deploy, incur lower costs, and require fewer resources, thereby contributing to both environmental and financial efficiency. Additionally, their reduced memorization and overfitting tendencies mitigate privacy and security risks. Furthermore, smaller models are often more interpretable, facilitating clearer insights into model decision-making processes.

\section{Conclusion}
\label{sec:conclusion}
In conclusion, our survey paper makes valuable contributions to the implementation of LLMs by providing an in-depth review of deployment risks, identifying underlying causes, and reviewing viable mitigation solutions. The study highlights various risks associated with LLM adoption, including ethical, privacy, security, bias, environmental, and compliance issues. We analyze factors contributing to these risks, such as model overfitting, complex architectures, limited awareness, lack of legislative uniformity in AI ethics, evolving threat landscapes, and insufficient control over data quality. To address these challenges, we emphasize proactive measures such as building robust models, employing privacy-preserving practices, implementing regulatory compliance measures, incorporating bias detection mechanisms, using explainability tools, and adopting parameter-efficient models. By applying these recommendations, researchers and stakeholders can advance the responsible development and deployment of LLMs, resulting in improved reliability, safeguarded user privacy, enhanced AI fairness, and reduced environmental impacts.

\bibliographystyle{abbrv}
\bibliography{main} 

\end{document}